\documentclass[preprint,12pt]{elsarticle}

\usepackage{amssymb}
\usepackage{amsmath}
\usepackage{booktabs}
\usepackage{multirow}
\usepackage{tabularx}
\usepackage{graphicx}
\usepackage{array}
\usepackage{adjustbox}
\usepackage{makecell}
\usepackage{hyperref}
\usepackage{color}
\usepackage{dashrule}
\usepackage{color}
\usepackage{tabularray}
\usepackage{arydshln}
\usepackage{ragged2e}
\usepackage{ulem}
\usepackage[thicklines]{cancel}
\definecolor{revision}{RGB}{0,0,0}

\journal{Arxiv}

\begin{document}

\begin{frontmatter}



\title{\textcolor{revision}{NNPP: A Learning-Based Heuristic Model for Accelerating Optimal Path Planning on Uneven Terrain}}

\author[1]{Yiming Ji}
\ead{yimingji_hit@163.com}
\author[2]{Yang Liu \corref{cor1}}
\ead{liuyanghit@hit.edu.cn}
\cortext[cor1]{Corresponding author}
\author[3]{Guanghu Xie}
\author[4]{\textcolor{revision}{Boyu Ma}}
\author[5]{Zongwu Xie}
\author[9]{Baoshi Cao}
\affiliation[1,2,3,4,5,6]{organization={State Key Laboratory of Robotics and System},
            addressline={Harbin Institute of Technology},
            city={Harbin},
            postcode={150001},
            state={Heilongjiang},
            country={China}}

\begin{abstract}
\textcolor{revision}{Intelligent autonomous path planning is essential for enhancing the exploration efficiency of mobile robots operating in uneven terrains like planetary surfaces and off-road environments.}
\textcolor{revision}{In this paper, we propose the NNPP model for computing the heuristic region, enabling foundation algorithms like ${A^\star}$ to find the optimal path solely within this reduced search space, effectively decreasing the search time.}
The NNPP model learns semantic information about start and goal locations, as well as map representations, from numerous pre-annotated optimal path demonstrations, and produces a probabilistic distribution over each pixel representing the likelihood of it belonging to an optimal path on the map.
More specifically, the paper computes the traversal cost for each grid cell from the slope, roughness and elevation difference obtained from the \textcolor{revision}{digital elevation model}. Subsequently, the start and goal locations are encoded using a Gaussian distribution and different location encoding parameters are analyzed for their effect on model performance.
After training, the NNPP model is able to  \textcolor{revision}{accelerate} path planning on novel maps. 
\textcolor{revision}
{Experiments demonstrate that the heuristic region generated by the NNPP model achieves a 3$\times$speedup for optimal path planning under identical hardware conditions. Moreover, the NNPP model's advantage becomes more pronounced as the size of the map increases.}
\end{abstract}


\end{frontmatter}
\section{INTRODUCTION}
\textcolor{revision}{Path planning is the process of finding a desired path between two given locations while satisfying predefined requirements. With the emergence of field mobile robotics, such as rescue robots \cite{li2024novel} and planetary rovers \cite{sutoh2015right}, path planning in outdoor environments has gained significant attention recently.
Outdoor unstructured environments often feature uneven terrains where some paths are too hazardous for mobile robots to traverse due to instability on steep slopes and limited motion power \cite{yuan2017consistent}. Furthermore, unlike indoor scenes, the shortest paths in uneven terrains are not always optimal for mobile platforms as they often have to deal with rapid elevation changes. When mapping routes through rugged terrains, it's essential to account for more than just obstacles such as jutting rocks and sudden drops. It is also necessary to strike a balance between the shortest geometric distance and the potential dangers posed by uneven terrain. This complexity often stumps conventional algorithms like ${A^\star}$ and RRT, either leading to suboptimal paths or excessive computation time \cite{ganganath2018shortest}. Therefore, modeling terrain and conducting traversability analysis, as well as establishing criteria for evaluating the quality of a path, are essential prerequisites for path planning.}

To address the \textcolor{revision}{traversability analysis} problem of mobile robots in rugged terrains such as planet surfaces, researchers initially proposed the idea of binary terrain classification into impassable obstacles and free traversal regions \cite{laubach1998autonomous}. In  \cite{ettlin2006rough}, Ettlin suggested combining binary motion planners with traversability classification by integrating rapidly-exploring random trees and gradually increasing the threshold for binary terrain classification.
\textcolor{revision}{However, when navigating rough terrain, this binary categorization proves ineffective due to the continuous spectrum of terrain complexities since classifying an obstacle as non-traversable can result in a suboptimal plan or even failure to find a path. 
Thus, an adequate model capturing terrain traversability is essential for a nuanced environment representation and informed navigation planning.
}



The \textcolor{revision}{traversability of the uneven terrain} depends on the scene information contained within the map. 
This information includes the layout of the terrain, the presence of obstacles, environmental conditions, and other relevant factors. 
\textcolor{revision}{For instances, in uneven planetary terrain,} map data that includes height information can better reflect the requirements of planetary rovers for terrain exploration.
\begin{figure}[htb]
    \centering
    \includegraphics[width=\textwidth]{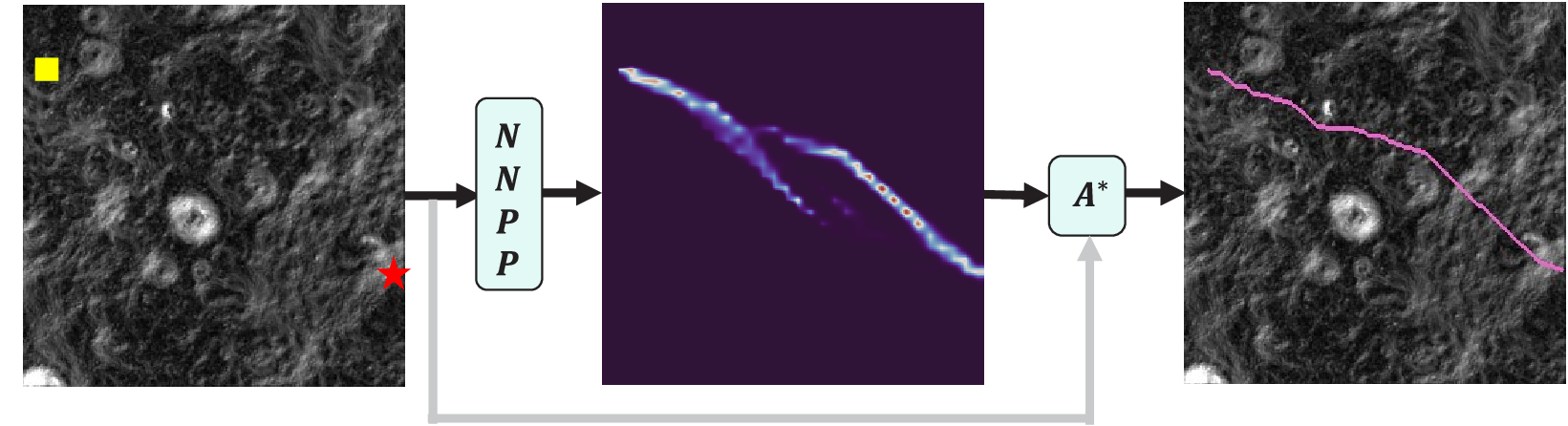}
    \caption{In the proposed framework, a cost map with start and goal points is fed into the NNPP. The model generates a probability map where each pixel's probability value indicates the likelihood of the optimal path passing through that pixel. This probability map is then utilized to guide the ${A^\star}$ algorithm on the original map, resulting in the final optimal path.}
    \label{fig:re_eps1}
\end{figure}
Thus, this paper adopts DEM data as the \textcolor{revision}{uneven} map representation and formulates the global path planning problem as finding the optimal feasible path that avoids untraversable areas, including obstacles, and minimizes the risk of failure due to hazardous terrain morphology, as shown in Figure \ref{fig:re_eps1}. 

The remainder of the paper is organized as follows. Section 2 introduces related work on path planning, focusing on learning-based path planning methods. 
Section 3 \textcolor{revision}{presents environmental modeling and cost map calculation methods, and investigates the trade-off between minimizing total cost and minimizing total path length.}
Section 4 elaborates on the proposed method, including the network structure, dataset preparation, and model evaluation metrics. Section 5 introduces the simulation experiments designed for the model, which demonstrates that the model can significantly reduce optimal path search time.

\section{RELATED WORKS}

Numerous algorithms have been proposed and refined to tackle the path planning problem, including the ${A^\star}$ algorithm \cite{hart1968formal}, RRT-based algorithms \cite{lavalle2001randomized}, artificial potential field (APF) based algorithms \cite{niu2023virtual}, and intelligent algorithms such as the genetic algorithm \cite{zhang2016dynamic}, ant colony algorithm \cite{liu2017improved} and reinforcement learning based algorithms \cite{jaradat2011reinforcement}. This section focuses on the mainstream methods of path planning and their applications in rough terrain navigation.

\subsection{Heuristic method}
Heuristic path planning algorithms utilize a heuristic function ${H\left( n \right)}$ to select the extended nodes, significantly reducing the total number of extended nodes and improving the efficiency of trajectory planning \cite{wang2021warehouse}. In a Grid map environment, the ${A^\star}$ algorithm can effectively represent obstacles of arbitrary shapes in two-dimensional space. However, in a three-dimensional environment, the evaluation function becomes complicated with the increase in dimensionality. Thus, the ${A^\star}$ algorithm is less commonly used in high-dimensional space. To address this limitation, many studies have proposed optimization schemes based on ${A^\star}$ algorithms \cite{astar2020}, such as ${D^\star}$, ${Field D^\star}$, ${Theta^\star}$, Anytime Repairing ${A^\star}$ (${ARA^\star}$), and more.

In the elevation map scenario, Bo Zhang \cite{zhang2022path} utilized the cumulative traversability cost between grids as the cost function for ${A^\star}$ algorithm, thereby enabling it to search for the shortest path in 3D uneven terrains.

\subsection{Learning based methods}
In recent years, modulated by the rapid development and widespread applications of artificial intelligence technologies, especially the advances within deep reinforcement learning (DRL), learning-based methods for path planning, obstacle detection, trafficability analysis and related path reasoning tasks have received increasing interest from researchers in both academia and industry hoping to enable more autonomous navigation for unmanned systems operating in complex environments \cite{tai2017virtual} \cite{zhou2019vision}.

Based on the heuristic methods discussed in the previous section, many researchers have employed learning-based techniques to construct guiding functions that accelerate the convergence process of classical algorithms. This approach has proven to be highly effective. The ${NRRT^\star}$ \cite{wang2020neural} approach applies a convolutional neural network (CNN) model to guide the sampling-based RRT path search on a 2D grid map while satisfying different clearance and step size planning requirements. The same team of authors proposed the NEED model \cite{wang2021robot}, which predicts promising search regions to guide the search direction of planning algorithms. This approach significantly enhances algorithm performance on new problems while maintaining a prediction speed of over 100 Hz on a laptop for high-dynamics applications.
Reinforcement learning (RL), compared to traditional control methods, has the notable advantages of not requiring environmental maps, possessing strong learning capabilities from data, and exhibiting high dynamic adaptability to changes in complex environments \cite{moreira2020deep}. Its deep neural network architecture can effectively process high-dimensional information from various sensors, while its reinforcement learning mechanism can perform continuous decision-making tasks in complex and dynamic environments. This makes it particularly suitable for local path planning in dynamic scenarios. Many researchers have attempted to apply reinforcement learning methods to local path planning in rough terrain. Q-learning has been applied to local path planning, and according to Reference  \cite{zhang2022path}, reinforcement learning methods can yield superior paths compared to traditional approaches. Reference  \cite{josef2020deep} indicates that compared to baseline methods, reinforcement learning techniques can improve the success rate of path planning. It can also be inferred that reinforcement learning methods consume more time compared to traditional approaches.

The time taken by the Q-learning method is almost double that of the APF method and approximately 15 times that of the conventional ${A^\star}$ method. The success rate of reinforcement learning (95\%-96\%) also impacts its applicability to global maps at a larger scale. For example, Reference  \cite{zhang2022path} reports that the proposed reinforcement learning method consumes nearly 11 times the time compared to traditional ${A^\star}$ methods when searching paths in a ${110 \times 110}$ elevation map, while only improving path optimality by 20\%. The proposed NNPP method in this paper not only greatly reduces the search time (shortened by \textcolor{revision}{3} times) and algorithm optimality is nearly on par with ${A^\star}$ (reduced by 2\%) compared to traditional ${A^\star}$ on the same hardware and a ${256\times256}$ map, but NNPP's performance advantage will further increase with the expansion of the map size.

\section{ENVIRONMENTAL MODELING AND PROBLEM FORMULATION}
\label{sec:Environmental_Modeling}
\textcolor{revision}{The environmental modeling processes, such as grid-based, graph-based, and point-based methods, are prerequisites for path planning.}
Digital Elevation Model (DEM) is commonly used in many instances of path planning on uneven terrains, thus this chapter introduces DEM-based environmental modeling approaches as well as the problem formulation discussed in this paper.

\subsection{Digital Representation of Elevation Map}
In practical applications of CE4 \cite{wang2020vision}, DEM is used for map generation and path searching. The spatial distribution of an uneven terrain environment is characterized utilizing a group of three-dimensional vectors ${\left\{ {x,y,z} \right\}}$ in which ${x}$ and ${y}$ represent the planar coordinates of mapping an environmental feature point within the uneven terrain while ${z}$ denotes the altitude of the point. Therefore, geomorphic attributes including aspect, gradient of slope, changes in elevation, and roughness can be adequately extracted via DEM.
\subsection{Information Extraction of Uneven Terrain Environment}

According to the traversability assessment method used by Yutu 2 lunar rover \cite{yan2019autonomous}, geomorphic attributes of the ${i{\rm{ - th}}}$ grid cell in DEM data are calculated by the Eq.\ref{eq:eq1}. To calculate the roughness, a patch consisting of the ${i{\rm{ - th}}}$ grid cell and its eight neighboring cells is used.

\begin{equation}
    \label{eq:eq1}
    {T_i} = {k_s} \cdot \frac{{{\varphi _i}}}{{{\varphi _s}}} + {k_r} \cdot \frac{{{r_i}}}{{{r_s}}} + {k_f} \cdot \frac{{\Delta h}}{{{H_s}}}
\end{equation}

\begin{figure}[htb]
    \centering
    \includegraphics[width=\textwidth]{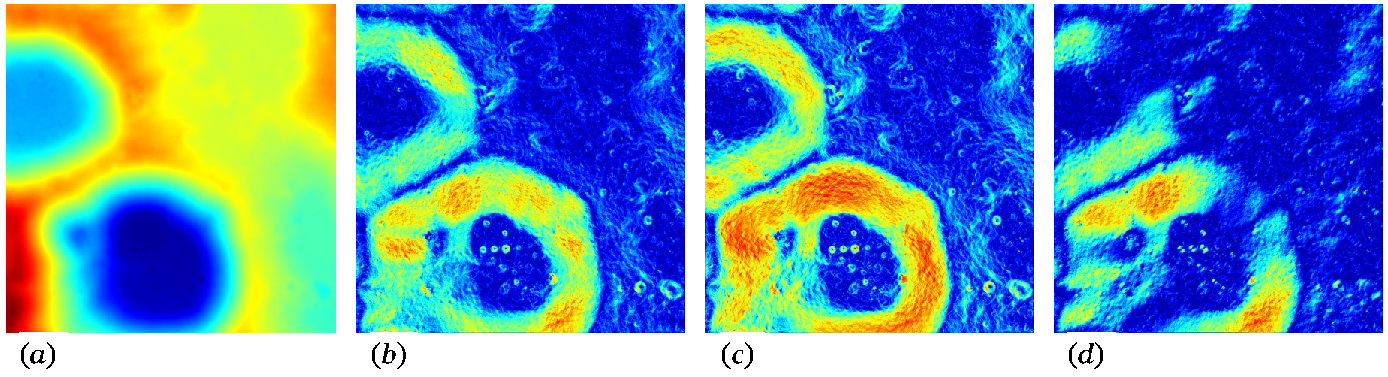}
    \caption{The traversal cost for each pixel was calculated using the following process: ${(a)}$ the original DEM was obtained; ${(b)}$ the slope cost was computed; ${(c)}$ the roughness cost was calculated; ${(d)}$ the elevation difference cost was determined. The resulting images, displayed from left to right, represent the original DEM, slope cost, roughness cost, and elevation difference cost. In this paper, we set ${\varphi_s = 30°}$, ${H_s = 0.2}$, ${r_s = 0.6}$.}
    \label{fig:re_eps2}
\end{figure}

The angle between the normal vector and horizontal plane of grid ${i}$ is defined as ${\varphi _i}$. The variable ${r}$ denotes the residual of the fitting distance from each DEM data point in a unit cell to the fitted plane in the corresponding patch, representing the roughness of the terrain region. The variable ${\Delta h}$ represents the height difference between each DEM data point in a unit cell and the fitted plane in the corresponding patch, representing the elevation difference in the terrain region.
The traversability cost for each grid cell is calculated as shown in Figure \ref{fig:re_eps2}.

In Eq.\ref{eq:eq1}, ${\varphi _s}$ represents the maximum safe tilt angle that a rover can traverse through a single grid cell, ${H _s}$ represents the maximum obstacle height value, and \textcolor{revision}{${r_s}$} represents the maximum roughness value that a rover can navigate through. The variables ${k _s}$, ${k _r}$, and ${k _f}$  are weighting coefficients, and their sum is equal to 1.
\textcolor{revision}{In this study, we set the parameters ${k_s}$, ${k_r}$, and ${k_f}$ to 0.6, 0.2, and 0.2, respectively.
Among these parameters, ${k_s}$ has the greatest impact on the robot's performance, as it accounts for the influence of terrain slope. Therefore, we assign a larger value to  ${k_s}$ compared to ${k_r}$ and ${k_f}$ \cite{ma2020geometry}.
The specific values of ${\varphi _s}$, ${H_s}$ and ${r_s}$ are determined by the robot's structural and kinematic-dynamic parameters.
As this paper focuses on planetary surfaces as a research case, we have referenced relevant data from lunar and Martian rovers \cite{zhang2022slip} to establish values for ${\varphi _s}$, ${H_s}$ and ${r_s}$  set respectively at 30 deg, 0.2 m, and 0.6.
}
Let the traversability cost for traveling between the two grids $i$ and \textcolor{revision}{${i+1}$ }be Eq.\ref{eq:eq2}.

\begin{equation}
    \label{eq:eq2}
    \textcolor{revision}{
        {T_{i,i+1}} = T_i + T_{i+1}
        }
\end{equation}
We adopt the approach of modeling the environment by calculating the cost of passing through each pixel value because in rugged and uneven terrains, the shortest Euclidean path may be impassable, thus it is necessary to integrally describe the terrain environment by combining slope, roughness and other geometric features in order to find the optimal traversable path. 
After the aforementioned environmental modeling, the problem discussed in this paper can be abstracted as \textcolor{revision}{searching for a path from a given start point to the end in a 2.5D cost map that
strikes an ideal balance between accumulated traversability cost and Euclidean path length.}

\begin{figure}[htb] 
    \textcolor{revision}{
    \centering
    \includegraphics[width=10 cm]{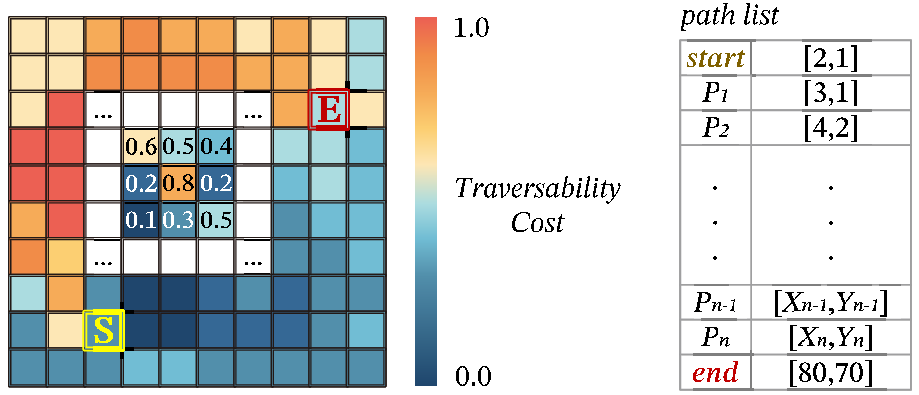}
    \caption{Path planning task, including a traversability cost map illustration and a path illustration}
    \label{fig:re_eps3}}
\end{figure}

\textcolor{revision}{
\subsection{Task definition}}
\label{sec:Task_definition}
\textcolor{revision}{
As described in previous sections, the DEM of the uneven terrain is transformed into a traversability cost map as shown in Figure \ref{fig:re_eps3}. Each grid cell in the map is assigned a cost value between 0 and 1.
A path is composed of a sequence of adjacent grid coordinates. The Euclidean length of a path is the sum of the Euclidean distances between adjacent grid cells along the path, as shown in Eq.\ref{eq:EuclideanLen}. The cumulative traverse cost of a path represents the sum of the traverse cost values of the grid cells ($T_i$) along the path, denoted as $CC$, as Eq.\ref{eq:cc}.
}

\begin{equation}
    \label{eq:EuclideanLen} 
    \textcolor{revision}{
    {L = \sum {_{i=1}^{n-1}}{\sqrt{(X_{i+1}-X_i)^2 + (Y_{i+1}-Y_i)^2}}}}
\end{equation}

\begin{equation}
    \label{eq:cc} 
    \textcolor{revision}{
        {{CC} = \sum{_{i=0}^n} {{T_i}}}}
\end{equation}

\textcolor{revision}
{
In path planning tasks, the input consists of a traversability cost map and the start and goal coordinates, and the model outputs a path represented by a sequence of grid coordinates. This path minimizes the weighted length (${L} + \omega  \cdot {CC}$).
The intricate relationship between the $\omega$ and the balance between $CC$ and $L$ is delved into in Section \ref{sec:Dataset_Generation}, leading to the identification of an optimal value of $\omega$.
}
\section{NNPP ALGORITHM}
\subsection{Network Structure}
In this section, \textcolor{revision}{we introduce the neural network structure of the proposed NNPP model}. The model takes as input a three-channel combination tensor, where the first channel represents a cost map as detailed in \textcolor{revision}{Section \ref{sec:Environmental_Modeling}}, with the grayscale value of each pixel indicating the cost of the robot passing through that pixel, with higher costs indicating pixels that are closer to being impassable. The second channel encodes the path planning starting point using a Gaussian representation, as described in \textcolor{revision}{Section \ref{sec:Training_Process}}. Similarly, the third channel represents the endpoint's Gaussian encoding. The model's output is a two-dimensional single-channel probability map, visualized as a color map in Figure \ref{fig:eps2}. Each pixel in the probability map stores a probability value ${p}$ between 0 and 1, indicating the probability that the pixel contains the optimal path. Higher probabilities imply the model is more inclined to believe the optimal path passes through that pixel.

\begin{figure}[htb]
    \centering
    \includegraphics[width=\textwidth]{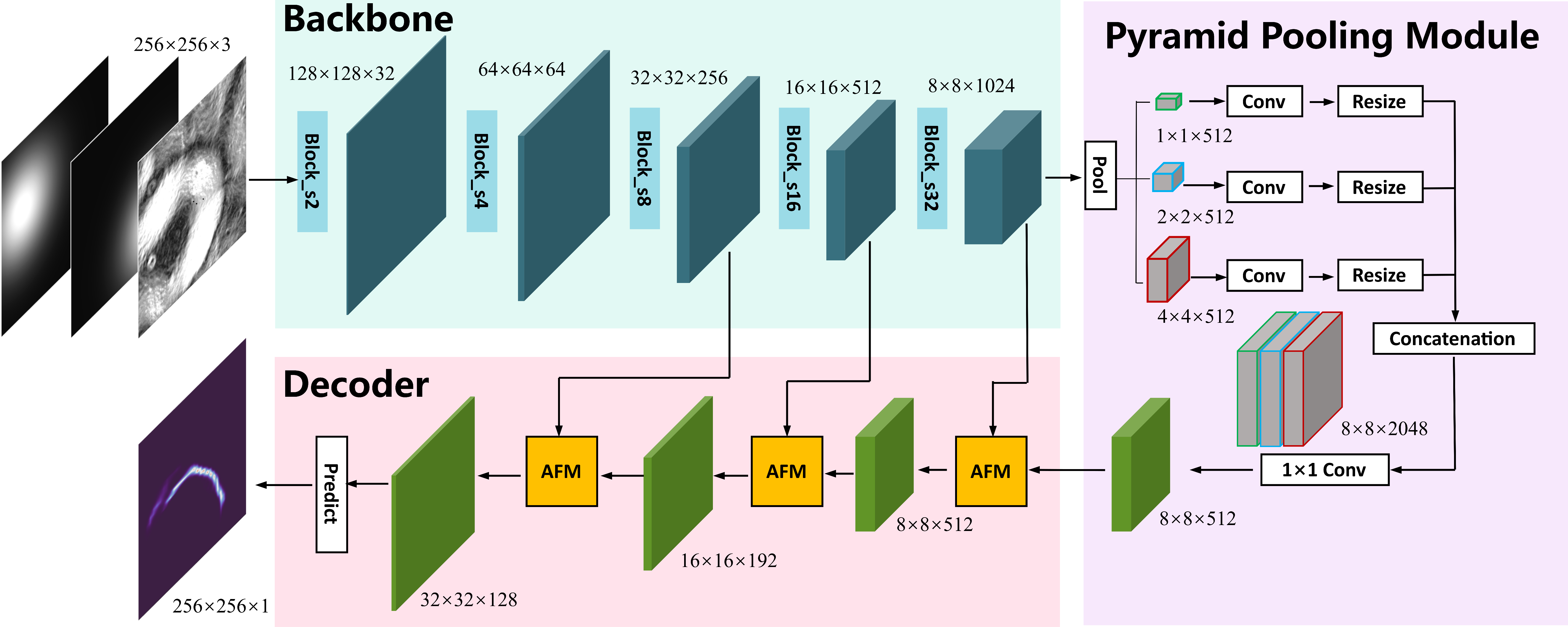}
    \caption{The model architecture diagram. It consists of an encoder, a decoder, and a contextual structure. The input is a three-channel tensor, and the output is a single-channel tensor.}
    \label{fig:re_eps4}
\end{figure}

The detailed structure of the model is shown in Figure \ref{fig:re_eps4}. It uses an encoder-decoder architecture comprising a backbone module, pyramid pooling module and decoder module. From a unique perspective, the role of the model is equivalent to performing semantic segmentation on the map containing the start and end points, segmenting out those pixels that represent feasible paths, while we also expect the model to have a very fast inference speed. Therefore, in designing the model, we drew inspiration from excellent examples of real-time semantic segmentation models.

The NNPP model employs the STDC \cite{fan2021rethinking} as the backbone network. Within its multi-stage encoding architecture, the spatial resolution of feature maps is reduced by a factor of two after each STDC stage. As such, a high-level semantic description of the input map can be extracted as the final \textcolor{revision}{stage5} output of the STDC backbone. The low-level features output by \textcolor{revision}{stage3 and stage4} will be fused with the high-level features in the decoder to enhance the model's performance for path planning. Feature fusion across different semantic levels has been shown to be effective in improving performance for tasks requiring precise localization such as path planning \cite{wang2020neural}, by incorporating both low-level spatial details and high-level semantic information. Therefore, in our model's decoder, we fuse the low-level features from earlier stages with the high-level features, in order to leverage spatial details that can help produce more accurate path planning outputs.

To enhance global scene contextualization, PSPNet \cite{zhao2017pyramid} incorporated a pyramid pooling module (PPM) that concatenates pooling representations at multiple scales prior to the convolution layers, thereby forming both local component features and global contextual representations. Compared to the original PPM, we simplify it to use three different pyramid scales, and remove the concatenate operation between the original feature map and the upsampled features, as in the original PPM. Our simplified PPM can reduce computational cost. Specifically, we reduce the number of pyramid scales from the original 4 to only 3. Previous work has shown that features at different pyramid levels become increasingly semantic \cite{fu2019dual}. Therefore, a fewer number of pyramid levels, such as 3, can capture sufficient semantic information, while reducing computational cost.

\begin{figure}[htb]
    \centering
    \includegraphics[width=10 cm]{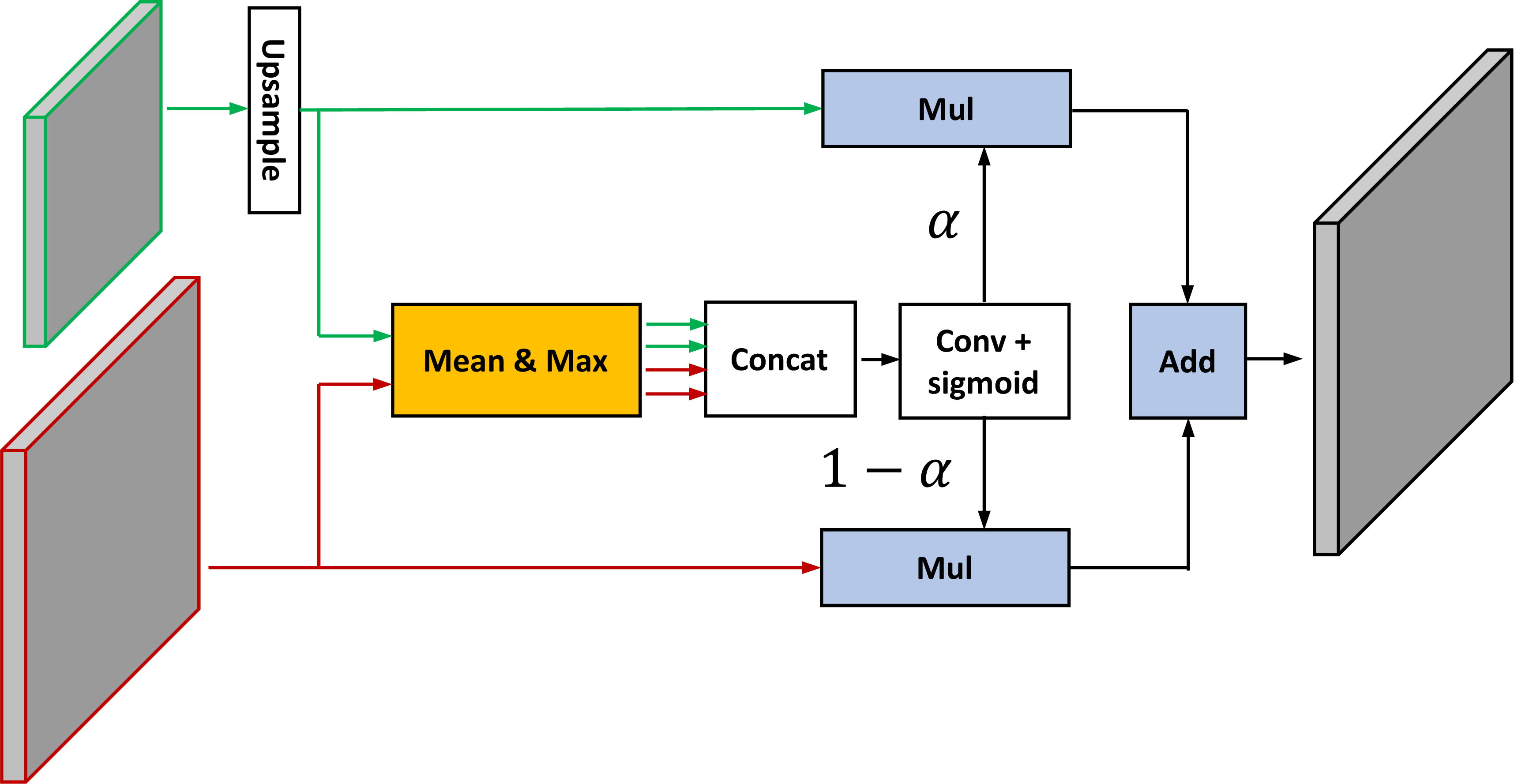}
    \caption{Attention fusing model. The model fuses the low-level features from earlier stages with the high-level features, in order to leverage spatial details that can help produce more accurate path planning outputs.}
    \label{fig:re_eps5}
\end{figure}

Feature fusion modules are commonly employed in semantic segmentation networks to enhance feature representations. More specifically, fusing multi-level features is instrumental in achieving high segmentation accuracy. The proposed model draws inspiration from the design of ppliteseg \cite{peng2022pp}, utilizing an Attention Fusion Module to amalgamate low-level features and high-level features from the decoder.
The structure of the Attention Fusion Module (AFM) \cite{peng2022pp} is shown in Figure \ref{fig:re_eps5}. 
 
The decoder likewise comprises several stages, which are tasked with merging and upsampling features. At each stage, features of different levels are fused through the AFM module. As the problem can be viewed as a pixel-wise binary classification problem, a conventional convolution layer with only one output channels serves as the classifier, producing the corresponding probability of the pixel belonging to the foreground or background class.

\subsection{Dataset Generation}
\label{sec:Dataset_Generation}
\textcolor{revision}
{Obtaining the ground truth optimal path on uneven terrain is challenging as it involves semantic segmentation and terrain traversability analysis using intrinsic parameters of the mobile platform. For instance, generating annotated data and training labels for planetary surfaces can be a very time-consuming task \cite{dai2022segmarsvit}.}
To address this issue, NNPP does not require any manually labeled data. Instead, it uses the optimal path obtained by ${A^\star}$  as a label to train the model. The 20-meter resolution lunar DEM dataset from CE2TMap2015 \cite{xin2020geometric} was used to generate a ${256\times256}$ map. The cost of traversing each pixel was calculated using the method outlined in Eq.\ref{eq:eq1}.

The classical ${A^\star}$ algorithm is a graph search algorithm commonly utilized for global path planning \cite{xie20202}. It is capable of finding the shortest path for a robot to travel between two predetermined positions in a known static environment. 
\textcolor{revision}{In the classical ${A^\star}$ algorithm, as in Eq.\ref{eq:eq3}, ${F\left( n \right)}$ represents the cost value of the ${n}$-th node. During the search process of the ${A^\star}$ algorithm, the cost values of all neighbor nodes of the current node are calculated, and the node with the smallest value is inserted into the open list.
}
\begin{equation}
    \label{eq:eq3}
    F\left( n \right) = G\left( n \right) + H\left( n \right)
\end{equation}

To facilitate the \textcolor{revision}{mobile robot's} automatic avoidance of untraversable areas caused by hazard \textcolor{revision}{uneven} terrain, this paper defines the cost function of ${A^\star}$ to include the traversability cost as follows:

\begin{equation}
    \label{eq:eq4}\textcolor{revision}{
    G\left( n \right) = {L_{sn}} + \omega  \cdot {T_{sn}}
    }
\end{equation}

Where ${{L_{sn}}}$ represents the accumulated Euclidean distance that from the starting grid ${s}$ to the current grid ${n}$, \textcolor{revision}{calculated as in Eq.\ref{eq:EuclideanLen},} 
\textcolor{revision}{and ${T_{sn}}$ represent the accumulated traversability cost from ${s}$ to ${n}$, ${{T_{sn}} = \sum _{k = 1}^{n - 1}{T_{k,k + 1}}}$ and ${T_{k,k + 1}}$ is computed as in Eq.\ref{eq:eq2}}
This paper set ${H\left( n \right)}$ to be the Euclidean heuristic function.

\textcolor{revision}{
    Next, we conducted a comprehensive analysis and established an optimal value for ${\omega}$ in Eq.\ref{eq:eq4}.
    An optimal path should simultaneously satisfy two criteria: (1) minimizing the $CC$ value which is cumulative traverse cost and (2) minimizing the Euclidean path length.
    When ${\omega=0}$, the algorithm ignores the traversability cost of each grid cell and directly searches for the shortest path from the start point to the end point. In most cases, this shortest path is a straight line. The larger the value of ${\omega}$, the greater the weight of the traversability cost.
    }

\textcolor{revision}{
    To determine the impact of ${\omega}$ on the traversability cost and Euclidean distance cost, we conducted additional experiments as shown in Figure \ref{fig:r1}.
}

\begin{figure}[htb]
    \centering
    \includegraphics[width=12.5 cm]{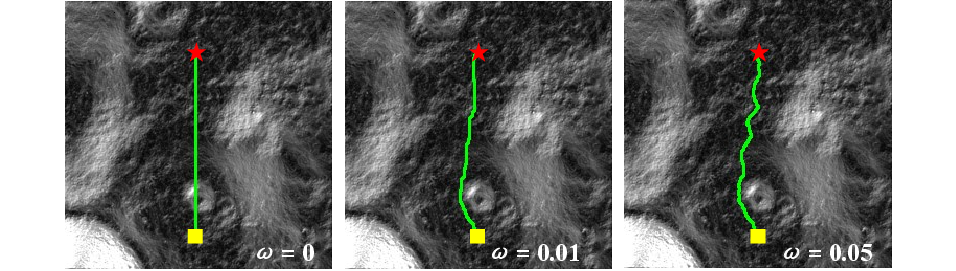}
    \textcolor{revision}{
    \caption{
        On the same map, with fixed start and end points, the path searching result using different weight values ${\omega}$. For better visualization, the path is shown in bold. The weight values ${\omega}$ from left to right are 0, 0.01, and 0.05.}
    \label{fig:re_eps6}
    }
\end{figure}
\begin{figure}[htb]
    \centering
    \includegraphics[width=12.5 cm]{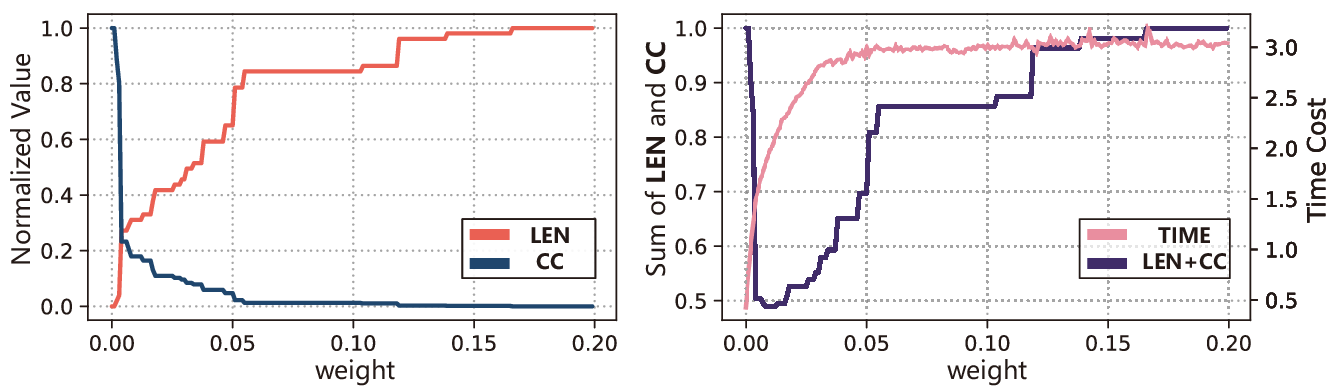}
    \textcolor{revision}{
    \caption{
        The left figure shows the relationship between ${\omega}$ and the normalized path length ($LEN$) and the normalized accumulated cost ($CC$), while the right figure is a double-y axis line graph. The left axis represents the change of $LEN+CC$ with ${\omega}$ (purple curve), and the right axis represents the time consumed for each path planning (pink curve).
         }
    \label{fig:re_eps7}
    }
\end{figure}

\textcolor{revision}
{As the value of ${\omega}$ increases, the path becomes more curved, as shown in the Figure \ref{fig:re_eps6}. This is to avoid high-cost grid cells and minimize the cumulative traverse cost of the path. 
Using the cost map in Figure \ref{fig:re_eps6} as an example, we fixed the start and end points and varied the ${\omega}$ from 0 to 0.2 to obtain the results shown in Figure \ref{fig:re_eps7}.
}

\textcolor{revision}
{Normalized accumulated Euclidean length and traversability cost of searched paths are denoted by $LEN$ and $CC$, respectively, as shown in the left subfigure of Figure \ref{fig:re_eps7}.
The two values were then added together to obtain the results shown in the right subfigure of Figure \ref{fig:re_eps7}. As the weight increases from zero, the algorithm returns a larger $LEN$ value and a smaller $CC$ value. The time consumed by the algorithm increases significantly at the beginning and then stabilizes.
}

\textcolor{revision}
{There exists a weight value that minimizes the sum of $LEN$ and $CC$ values, which we consider to be the optimal balance between path length and cumulative cost.
We selected 100 maps as samples, fixed the start and end points, calculated the optimal ${\omega}$ value for each sample, and obtained the result of ${0.011\pm0.003}$. When constructing subsequent datasets, we used ${{\omega} = 0.011}$ for all samples.
}
\subsection{Detailed Training Process and Evaluation Metric}
\label{sec:Training_Process}
Positional encoding is a powerful technique for infusing data with information regarding its position via embedding. In natural language processing (NLP), this approach is often leveraged to ensure that a model is aware of the location of words within a sentence. Inspired by this technique, NNPP makes use of positional encoding to embed location information about the starting and ending points on a map. The resulting encoded output is concatenated with the map, generating a three-channel tensor that serves as the input for the model. After encoding the starting point using sine and cosine functions to represent its absolute position \cite{vaswani2017attention}, the model's outputs were not feasible. However, by switching to relative position encoding, more desirable results were achieved. This is because in practical path planning, the system is less concerned with absolute positions and more focused on relative distances between points. Specifically, our interest lies in the relative position of each cell in the elevation map compared to the starting point and the goal point.
\begin{equation}
    \label{eq:eq5}
    g\left( {x,y} \right) = \frac{1}{{2\pi {\sigma ^2}}}{e^{ - \frac{{{{\left( {x - cx} \right)}^2} + {{\left( {y - cy} \right)}^2}}}{{2{\sigma ^2}}}}}
\end{equation}

In this paper, a two-dimensional Gaussian distribution function centered around the starting point is used as the position encoding, with each grid in the elevation map represented by its corresponding value in the Gaussian function. This value reflects the relative position of the grid with respect to the starting point. The same approach is taken for the ending point.
Eq.\ref{eq:eq5} represents the calculation method for the Gaussian distribution function value at ${(x, y)}$.
Parameter ${cx}$ represents the ${x}$-coordinate of either the starting or ending point, while ${cy}$ represents the ${y}$-coordinate.
\textcolor{revision}{The impact of the ${\sigma}$ parameter on model performance was analyzed in the experiments in section \ref{sec:Gaussian_Positional_Encoding}, and the optimal value was determined.}

In the field of path planning, scholars have proposed many evaluation methods for path quality \cite{liu2023path}, among which average path planning time is an important metric that reflects the search efficiency of the algorithm. In elevation maps, the proposed method uses the cumulative traverse cost of a path to measure its optimality, believing that an optimal path should have the minimum cumulative traverse cost.

\begin{figure}[htb]
    \centering
    \includegraphics[width=10 cm]{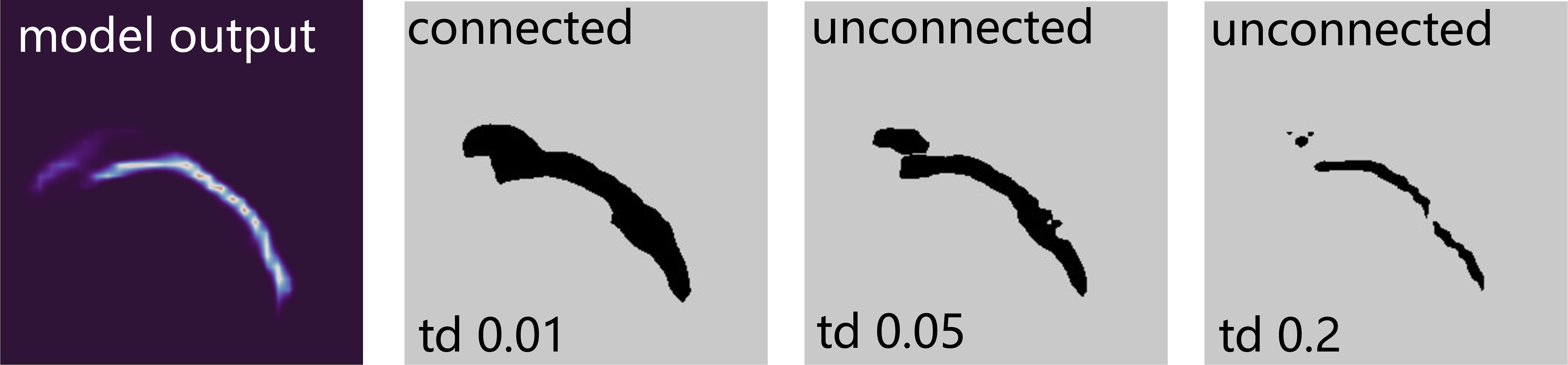}
    \caption{The segmentation results obtained with different threshold values ${(td)}$. The leftmost image represents the model's output map, while the remaining images depict the segmentation results obtained using different threshold values. The black regions indicate areas where the pixel values in the model's output map exceed the ${td}$.}
    \label{fig:re_eps8}
\end{figure}

This paper employs a learning-based approach to predict regions within a given map that contain optimal paths, whereas existing evaluation metrics are designed for a complete path rather than a region. Consequently, there is a need for designing an evaluation metric that directly assesses the precision of the model predictions. Figure \ref{fig:re_eps8} illustrates the results of segmenting the model outputs according to diverse thresholds ${td}$, with the black areas representing regions assigned probabilities lower than ${td}$ by the model. The designation ${'connected'}$ denotes whether the segmented regions unite the starting point and end point. The present study decreases ${td}$ gradually from high to low values while detecting whether the two points are connected, thereby determining an optimal binary threshold and the corresponding region segmentation map.

Subsequently, the total number of pixels that the model deems to contain optimal paths within the segmented regions is calculated, with the metric evaluated using the following Eq.\ref{eq:eq6}. \textcolor{revision}{${Z_{label}}$ represents the area of the labeled path, which is the sequence of grid cells obtained by running the ${A^\star}$ algorithm on the traversability cost map}, ${Z_{model}}$ denotes the area segmented using the adaptive threshold, and the areas are measured in number of pixels.
\textcolor{revision}
{
    $ModelMetric$, denoted as $MM$, measures the overlap between the heuristic region output by the NNPP and the ground truth optimal path. A higher $MM$ value indicates better model performance, while a lower value suggests poorer performance.}

\begin{equation}
    \label{eq:eq6}
    ModelMetric = \frac{{{Z_{label}}}}{{{Z_{model}}}}
\end{equation}


In summary, there are three evaluation metrics: the average path planning time ${(AT)}$, \textcolor{revision}{the cumulative traverse cost of a path (${CC = \sum {_i^{path}{T_i}}}$,${T_i}$ represent the traversability cost value of gird ${i}$ in the path grid list, as Eq.\ref{eq:eq1})}, and ${(MM)}$ as defined in Eq.\ref{eq:eq6}.

\section{SIMULATION EXPERIMENTS}
In \textcolor{revision}{this} section, we delve into intricate parameter tuning and hardware environment settings, designing three experiments to demonstrate the efficacy of the proposed method. As detailed in \ref{sec:Dataset_Generation}, a training set of 100,000 samples of resolution ${256\times256}$ was synthesized, 5000 samples were procured for evaluation, and an additional 1000 samples were sequestered for performance metric validation in the present experiments. The training, elevation and validation sets were mutually exclusive from each other. The optimizer of choice for parameter optimization was AdamW with an initial learning rate of 0.001 and a weight decay of ${{10^{ - 4}}}$. After training for 150 epochs, the model was able to predict traversable areas between start and end points on 1000 terrain samples that it had not been exposed to previously, showcasing its generalization capabilities.
\subsection{\textcolor{revision}{Comparative Experiments at Different Scales}}
The computationally intensive nature of utilizing algorithm ${A^\star}$ to search for optimal paths within elevation maps stems from the stipulation that ${A^\star}$ algorithm necessitates constructing a global search graph in advance of searching. 
\textcolor{revision}{
    The purpose of the graph is to store the open-list and closed-list during the search process and maintain the parent node of each node in the list. At each step of the search, the open-list is sorted to select the node with the highest current priority. Once the open-list grows to include the goal node, the parent nodes of the current node can be traced back from the graph to the starting node, thereby returning a path.
}

\begin{table}
    \centering
    \caption{Comprehensive experimental data on path planning using the ${A^\star}$ algorithm under different map resolutions. The path length measured in terms of the total number of pixels traversed along the path.}
    \label{tab:tab1}
    \resizebox{\textwidth}{!}{
    \begin{tabular}{>{\centering\hspace{0pt}}m{0.233\linewidth}>{\centering\hspace{0pt}}m{0.208\linewidth}>{\centering\hspace{0pt}}m{0.213\linewidth}>{\centering\hspace{0pt}}m{0.192\linewidth}>{\centering\arraybackslash\hspace{0pt}}m{0.152\linewidth}} 
    \hline\hline
    \textbf{Map Resolution} & \textbf{Graphing Time/s} & \textbf{Searching Time/s} & \textbf{Graphing Ratio} & \textbf{Path Length}  \\ 
    \hline
    $64\times64$            & 0.1854                   & 0.0230                    & 88.96\%                 & 57                    \\
    $128\times128$          & 0.6466                   & 0.1045                    & 86.09\%                 & 113                   \\
    $256\times256$          & 2.7770                   & 0.4715                    & 85.49\%                 & 226                   \\
    $512\times512$          & 11.1785                  & 2.0906                    & 84.24\%                 & 452                   \\
    $1024\times1024$        & 46.4525                  & 8.6014                    & 84.43\%                 & 904                   \\
    \hline\hline
    \end{tabular}}
    \end{table}
Table \ref{tab:tab1} presents the time consumption for pathfinding using ${A^\star}$ within maps of varying dimensions. The path length indicated in the table represents the quantity of pixels encompassed within the path as determined by ${A^\star}$. To more reasonably reflect the proportion of graphing time within the total search time, the same scaling operations are applied to the start and end points when the map is scaled proportionally.

Inspection reveals that graph construction accounts for approximately 85\% of the total time consumed by the algorithm. This entails an onerous outlay of computational resources necessitated by the construction and subsequent traversal of the exhaustive search graph. Indeed, the area subtended by optimal paths is markedly smaller than the total map area, indicating that a sizable proportion of the search graph constructed by ${A^\star}$ may be classified as unvaluable with respect to deriving optimal routes. In other words, the global search graph encompasses vast regions irrelevant to the final path, implying inefficiency in the algorithm's graph generation procedure.

\begin{figure}[htb]
    \centering
    \includegraphics[width=\textwidth]{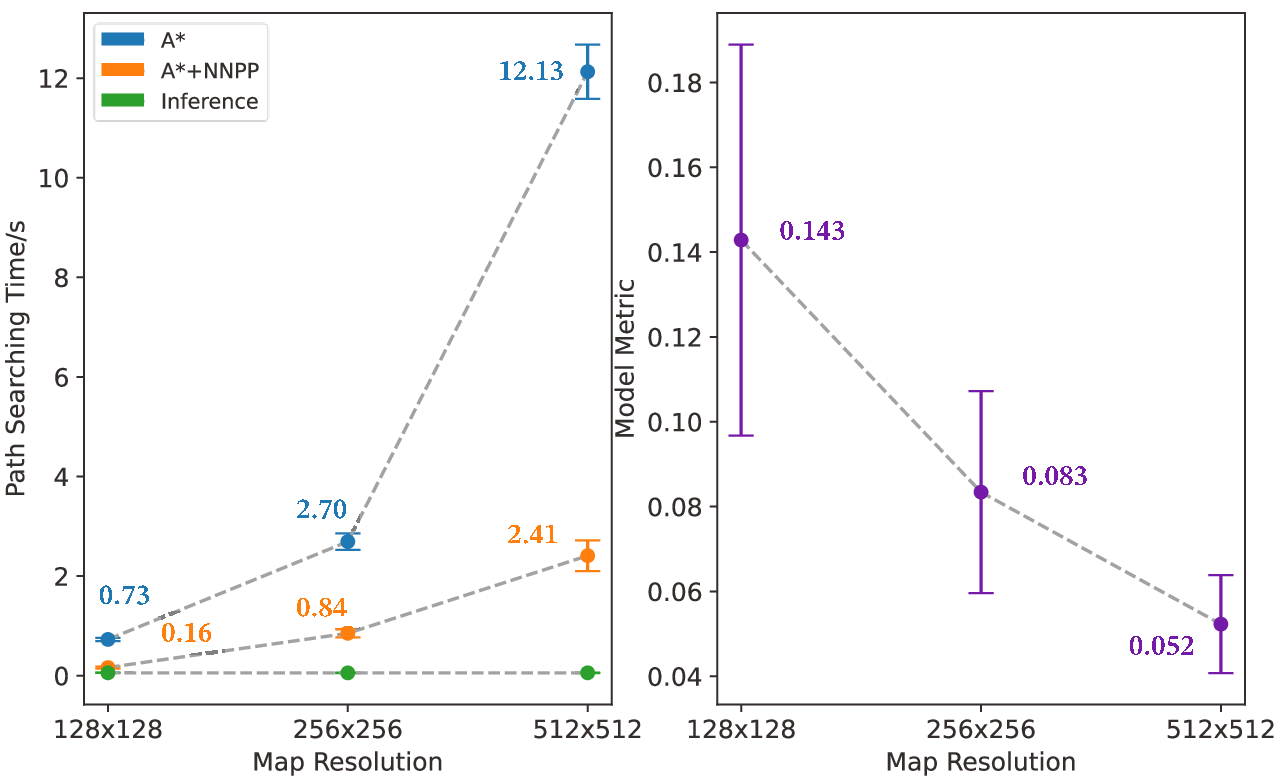}
    \caption{The left figure illustrates the variation of path search time with map resolution under different methods. The blue color represents the ${A^\star}$ search method, the orange color represents the method combining NNPP with ${A^\star}$, and the green color represents the model inference time. The model inference time remains nearly constant at 0.058s, indicating that the model can perform inference at a speed of 17 frames per second (FPS). On the right figure, the trend of Model Metric values for the NNPP method is shown as the map resolution changes.}
    \label{fig:re_eps9}
\end{figure}

However, the proposed \textcolor{revision}{NNPP} approach allows the neural network to focus on those regions that provide high value to subsequent path reconstruction algorithms such as ${RRT^\star}$ and ${A^\star}$. By reducing the spatial redundancy in the search space, the network can learn a more discriminative representation that facilitates more efficient replanning. 
Taking ${A^\star}$ as an example, constructing the search graph only within the \textcolor{revision}{heuristic region} output by NNPP, rather than a global search graph, as illustrated in Figure \ref{fig:re_eps9}, can save significant time. Moreover, as the map size increases, the advantages of the neural network approach become more pronounced. By leveraging the ability of the network to focus the search in the most promising areas, the proposed localized search graph yields increasingly superior speedup and scalability compared to exhaustive search-based algorithms as the problem size grows.

\begin{figure}[htb]
    \centering
    \includegraphics[width=12 cm]{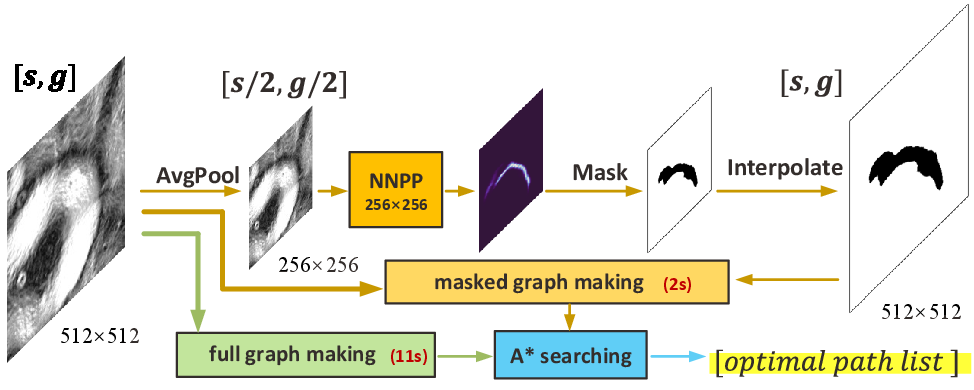}
    \caption{The computational approach of the model for maps of different scales. Here, taking a ${512\times512}$ map as an example, the ${"masked\;graph\;making"}$ approach guided by NNPP significantly reduces computation time compared to the traditional ${"full\;graph\;making"}$ method. }
    \label{fig:re_eps10}
\end{figure}

The model was trained on data of size ${256\times256}$. When evaluating the model's performance on ${512\times512}$ maps, as illustrated in Figure \ref{fig:re_eps10}, the maps were first downsampled via average pooling. Denoted by $s$ and $g$ are the start and goal positions respectively. Subsequently, the method described in \ref{sec:Training_Process} was applied to obtain the mask of the \textcolor{revision}{heuristic region} to the model. Finally, these regions were resized back to the original map size via interpolation. This procedure allowed the model trained on smaller maps to generalize well to larger maps, by focusing its search only on the relevant areas while ignoring the rest. The ability to downsample the input during inference and selectively pay attention to important regions enabled by the proposed method was critical for achieving scalability. The search graph was constructed only within the black regions in Figure \ref{fig:re_eps10}. Subsequently, the ${A^\star}$ method was applied on the search graph to compute the start-to-goal path. 
In comparison, an exhaustive search graph was built on the original map, and the same ${A^\star}$ method was utilized to search for the optimal path. \textcolor{revision}{As shown in Figure \ref{fig:re_eps15}, the heuristic region in the first column is notably smaller than the ${A^\star}$ algorithm search area in the second column.}
By constraining the search space to \textcolor{revision}{heuristic region}, the localized search graph approach achieved significantly higher efficiency and scalability compared to the exhaustive search baseline. The ability to prune irrelevant regions of the map based on the predicted importance masks allowed the search complexity to grow sublinearly with the map size, in contrast to the quadratic complexity of exhaustive search. This demonstrates the effectiveness of the proposed approach in focusing the search in the most promising regions to guide the path planning algorithm.

As shown in the Figure \ref{fig:re_eps10}, the masked graph making module accepts two inputs: the model output post masking and the original map. In the original map, regions of interest as discerned by the model are retained and delineated in black while the remainder is rendered entirely white, with the white regions signifying impassable obstacles. In this manner, ${A^\star}$ search for the optimal path is constrained to a relatively smaller region. Provided that the specified region encompasses the optimum path, ${A^\star}$ will return a 100\% consistent path matching the given label.

\begin{figure}[htb]
    \centering
    \includegraphics[width=10 cm]{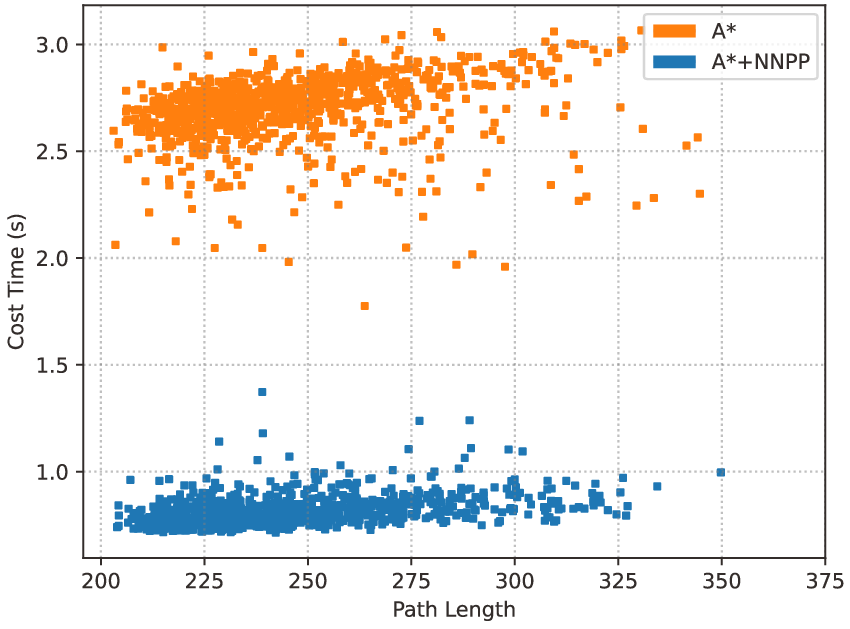}
    \caption{The relationship between the path length obtained by the algorithm and the search time is examined in the validation set. The blue scattered points represent the method combining NNPP with ${A^\star}$, while the orange color denotes the traditional ${A^\star}$ method. The fitted lines for both methods are also depicted in the figure. It can be observed that as the path length increases, the search time for that path also increases. The blue method exhibits significant superiority over the orange method. }
    \label{fig:re_eps11}
\end{figure}

As Figure \ref{fig:re_eps11} illustrates, the blue bars represent the computational time incurred by the NNPP method via the workflow in Figure \ref{fig:re_eps10} to derive the optimal path whereas the orange bars denote the time consumed by conventional ${A^\star}$ search directly on the map. Evidently, as the map size increases, the neural network more decisively shortens the search time. The right sub-figure of Figure \ref{fig:re_eps9} indicates the performance metrics measured using Eq.\ref{eq:eq6} and corresponding standard deviations for the NNPP model at diverse resolutions. 
In larger scenes, employing the upscaling via NNPP methodology to yield outputs compromises considerable original map information, resulting in the degraded \textcolor{revision}{$MM$ value}. 
\textcolor{revision}{
Yet, despite this information loss, the neural network approach can still remarkably shorten the time required to search for the optimal path. 
}

As shown in Table \ref{tab:tab2}, we tested the model's path planning performance on a validation set consisting of 1000 maps, wherein ${A^\star NN}$ denotes the approach using the procedure of Figure \ref{fig:re_eps10}, utilizing the model output to guide ${A^\star}$ for path replanning.

\begin{figure}[htb]
    \centering
    \includegraphics[width=10 cm]{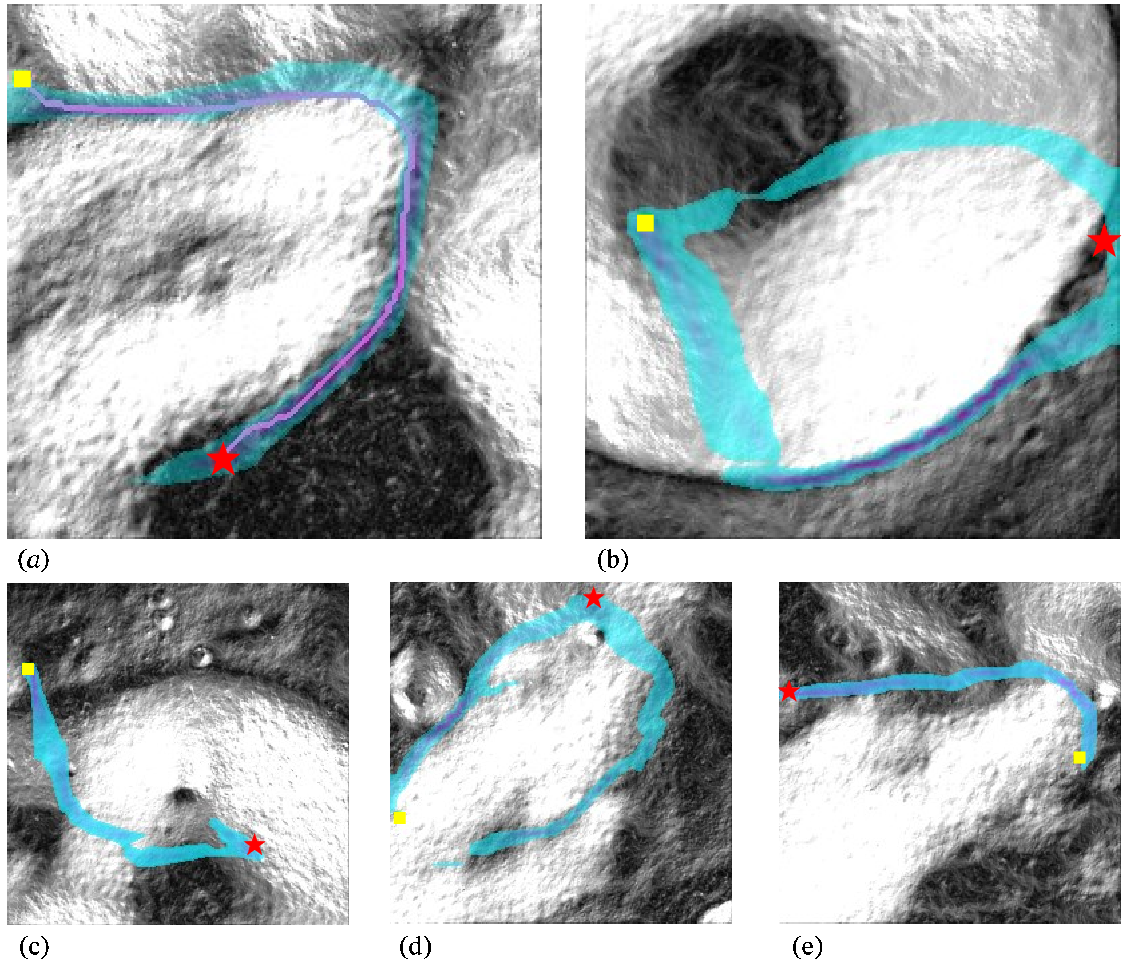}
    \caption{Some examples of path planning on a ${256\times256}$ map, where the high probability region, indicated by the color cyan, corresponds to the optimal path suggested by the model. The yellow square represents the starting point, while the red pentagram represents the destination.}
    \label{fig:re_eps12}
\end{figure} 

Figure \ref{fig:re_eps12} illustrates several scenarios of path planning on a ${256\times256}$ map. It can be observed that the model's output effectively avoids hazard terrain (where whiter pixels indicate higher traversal costs, indicating the increased steepness). The model provides relatively safe regions.
For certain scenarios, the model presents two alternative paths, indicating its ability to handle semantic information of the global map. During actual planning, the model selects the path with a higher probability value. The corresponding data for the five scenarios are presented in Table \ref{tab:tab3}.

\begin{table}
    \centering
    \caption{Detailed experimental data is provided for maps with resolutions of ${256\times256}$ and ${512\times512}$. In the table, ${AT}$ represents the average path search time, while ${CC}$ represents the cumulative cost of the path in the cost map.}
    \label{tab:tab2}
    \textcolor{revision}{
    \resizebox{\textwidth}{!}{
    \begin{tabular}{>{\centering\hspace{0pt}}m{0.223\linewidth}|>{\centering\hspace{0pt}}m{0.162\linewidth}>{\centering\hspace{0pt}}m{0.198\linewidth}|>{\centering\hspace{0pt}}m{0.162\linewidth}>{\centering\arraybackslash\hspace{0pt}}m{0.198\linewidth}} 
    \hline\hline
    \multirow{2}{1\linewidth}{\hspace{0pt}\Centering{}\textbf{metric}} & \multicolumn{2}{>{\Centering\hspace{0pt}}m{0.36\linewidth}|}{\textbf{${256\times256}$}} & \multicolumn{2}{>{\Centering\hspace{0pt}}m{0.36\linewidth}}{\textbf{${512\times512}$}}  \\ 
    \cline{2-5}
                                                                           & $A^\star$ & $A^\star NN$                                                                & $A^\star$ & $A^\star NN$                                                                \\ 
    \hline
    $AT$                                                                   & 2.7234    & 0.8511                                                                      & 12.1285   & 2.4103                                                                      \\
    $CC$                                                                   & 43.5157   & 44.8051                                                                     & 89.1457   & 91.2667                                                                     \\
    $Success Rate$                                                         & 100\%     & 99\%                                                                        & 100\%     & 99.4\%                                                                      \\
    \hline\hline
    \end{tabular}}}
    \end{table}


The data presented in the Table \ref{tab:tab2} substantiates the claim that the method proposed in this paper significantly reduces the time consumed in optimal path planning, without substantial compromise on the path's quality. Specifically, on a ${256\times256}$ grid map, the NNPP method has achieved a \textcolor{revision}{3.2}-fold acceleration in the path searching time of the ${A^\star}$ algorithm, with a mere 2.9\% decrease in path quality. As the map size increases, the advantages of the NNPP method are expected to become even more pronounced.

\textcolor{revision}
{
    There are many other methods, such as octrees and binary heaps, that can also effectively enhance the efficiency of graphing in ${A^\star}$. The NNPP model proposed in this paper does not enhance efficiency by altering the principles of graphing, instead, it narrows the scope of graphing by learning the distribution characteristics of optimal paths in the traversability cost map.
    Regardless of the graphing method employed, NNPP significantly improves algorithmic search efficiency.
}

\subsection{\textcolor{revision}{The Effects of Gaussian Positional Encoding}}
\label{sec:Gaussian_Positional_Encoding}
\textcolor{revision}{Unlike previous approaches, we input start and end point information, encoded using positional encoding, into the model via distinct channels.}
This approach effectively ensures that the model accords sufficient weight to the locational information of the start and end points during prediction. Literature \cite{kulvicius2021one} also inputs the start and end points via separate channels, but only with a single pixel each. While the approach can effectively predict paths in small grid maps up to ${20\times20}$, it struggles in larger and more complex scenes such as the elevation maps discussed herein. The experiments depicted in Figure \ref{fig:eps3} demonstrate that appropriate positional encoding of the start and end point parameters can considerably enhance the model's delineation of traversable areas, whereas utilizing single pixels without positional encoding leads to unsatisfactory performance.
\begin{figure}[htb]
    \centering
    \includegraphics[width=10 cm]{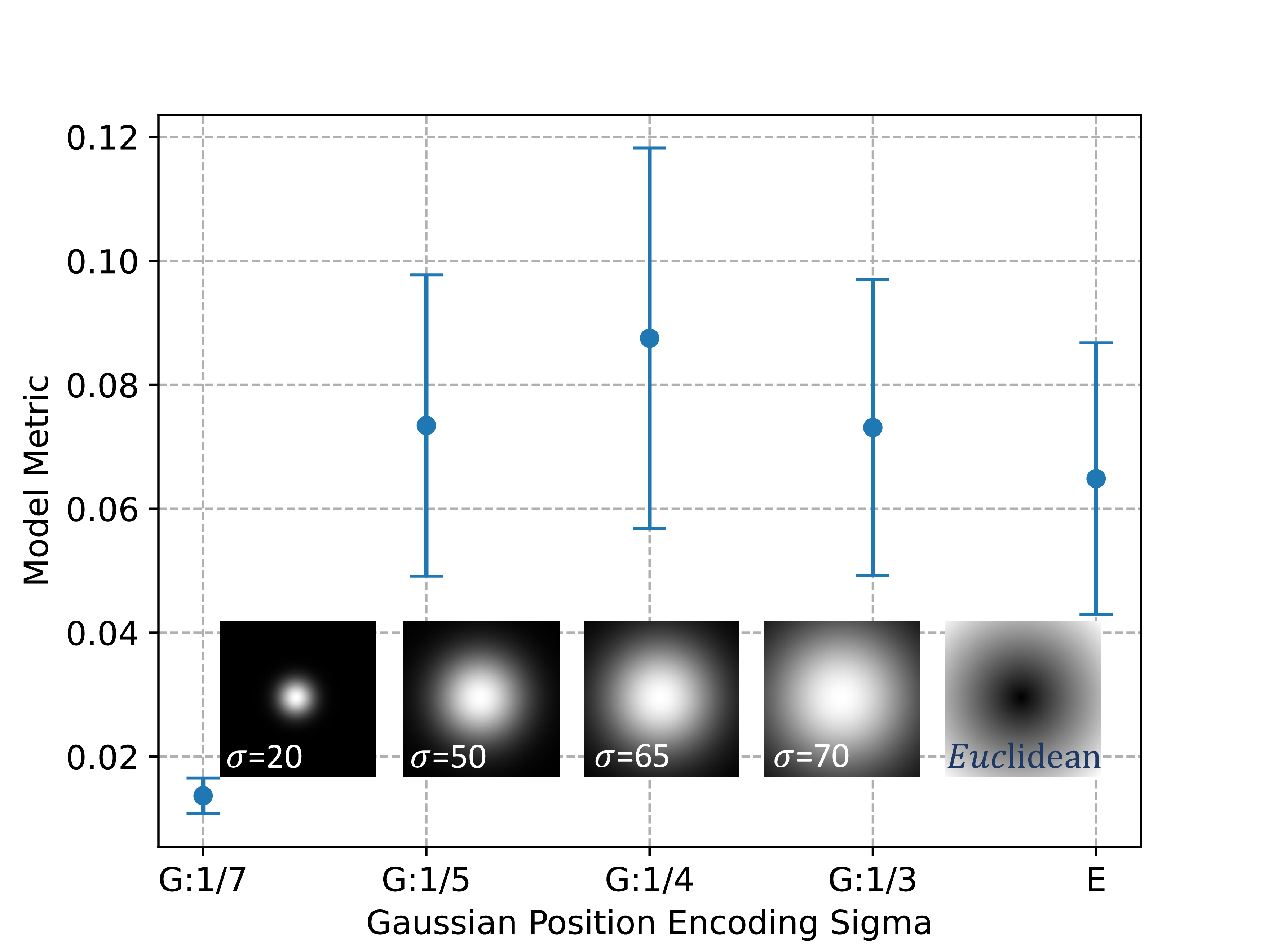}
    \caption{The impact of different positional encoding methods on model performance. In the horizontal axis, ${G}$ represents ${Gaussian}$ positional encoding, ${1/7}$ denotes the ${\sigma}$ value equal to one-seventh of the map size, and ${E}$ indicates ${Euclidean}$ encoding.}
    \label{fig:re_eps13}
\end{figure}
In Figure \ref{fig:re_eps13}, ${G}$ represents ${Gaussian}$ position encoding and ${E}$ represents ${Euclidean}$ position encoding. ${G: 1/7}$ indicates that ${\sigma}$ is taken as ${1/7}$ of the map size in Eq.\ref{eq:eq5}. The Euclidean encoding method calculates the Euclidean distance from each grid cell to the start or end point as the relative position encoding, which is then normalized to generate the final 2D encoding. It can be seen that the model prediction performs the best when ${\sigma}$ is around ${1/4}$ of the map size. When ${\sigma}$ is very small, it is close to no position encoding, resulting in a significant decrease in model prediction performance.


\subsection{Application to Dynamic Environment}
\label{sec:Dynamic_Environment}
\begin{figure}[htb]
    \centering
    \includegraphics[width=\textwidth]{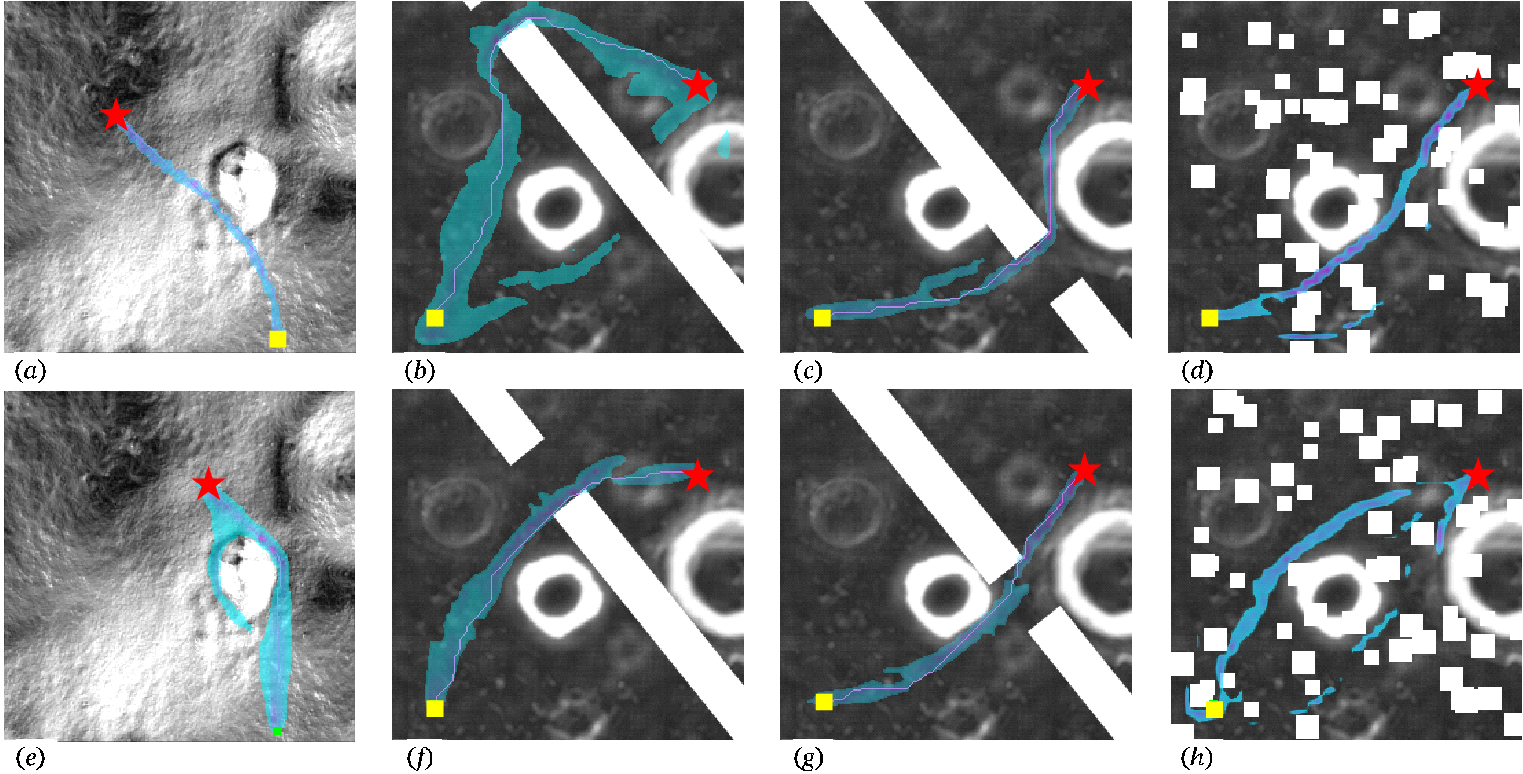}
    \caption{The results of model-based path planning in dynamic scenarios are presented, including scenarios with different endpoints for the same starting point ${(a)}$ and ${(e)}$, as well as scenarios with the same starting and ending points but changing maps ${(b)-(d)}$ and ${(f)-(h)}$.}
    \label{fig:re_eps14}
\end{figure}

For autonomous navigation robots, global path planning and local path planning target different scenarios, which are based on the global map from external sensors and local map from onboard sensors, respectively. 
\textcolor{revision}{Both global and local traversability cost maps derived from DEM data share the same format (${H \times W \times 1}$). However, they differ in two key aspects: 
\begin{itemize}
    \item (1) The real-world resolution corresponding to each grid cell. Figure \ref{fig:re_eps12} depicts a global map with a resolution of 20 meters, while the local map in Figure \ref{fig:re_eps14} has a resolution of 0.2 meters. 
    \item (2) The presence of occlusion artifacts in local maps due to the limited field of view of onboard sensors.
\end{itemize}   
We argue that, the NNPP model effectively addresses the task definition challenges outlined in Section \ref{sec:Task_definition}, regardless of whether the task is global or local.
}
The NNPP method introduced in this paper essentially learns the mapping relationship between the terrain fluctuations within a certain area measured by traversability cost and the optimal path given start and end points. Therefore, whether it is a global map or a local map, the NNPP is capable of predicting optimal path regions from uneven terrains. To validate this idea, in this section, we employ the NNPP trained on a global lunar map dataset for path planning in local Martian maps. To further simulate the presence of obstacles in the local maps, several artificial noise points are introduced into the local Martian Digital Elevation Model (DEM), as illustrated in Figure \ref{fig:re_eps14}.

Figure \ref{fig:re_eps14} ${(a)}$ and ${(e)}$ validate the path planning for reaching different destination points under the same scene and starting point, while Figure \ref{fig:re_eps14} ${(b)-(d)}$ and ${(f)-(h)}$ validate the path planning for the same starting and destination points in varying scenes.
\textcolor{revision}{Highlighting the adaptability of NNPP to dynamic environments due to its fast speed (17FPS), we further demonstrate its capabilities through additional experiments involving dynamic environments. Specifically, we introduce two obstacle regions into the traversability cost map, each assigned a traversal cost of 1 (hereafter referred to as obstacles).
}

\textcolor{revision}{
    As illustrated by the Figure \ref{fig:re_eps15} below, the NNPP model exhibits spatial consistency in its predictions. Despite the varying obstacle positions in timestamps $T$ and $T + 4\Delta t$, the NNPP model generates identical segmentation regions.
    This indicates that the NNPP model can effectively distinguish whether the location of obstacles influences the optimal path. In the demonstration scenario, the upper obstacle moves to the left, and the lower obstacle moves to the right. With the start and end points remaining unchanged, the NNPP model accurately estimates the optimal path distribution for each frame and then combines it with the ${A^\star}$ algorithm to determine the optimal path for the current frame. This demonstrates the NNPP model's ability to handle path planning in dynamic environments.
}
\begin{figure}[htb]
    \textcolor{revision}{
    \centering
    \includegraphics[width=\textwidth]{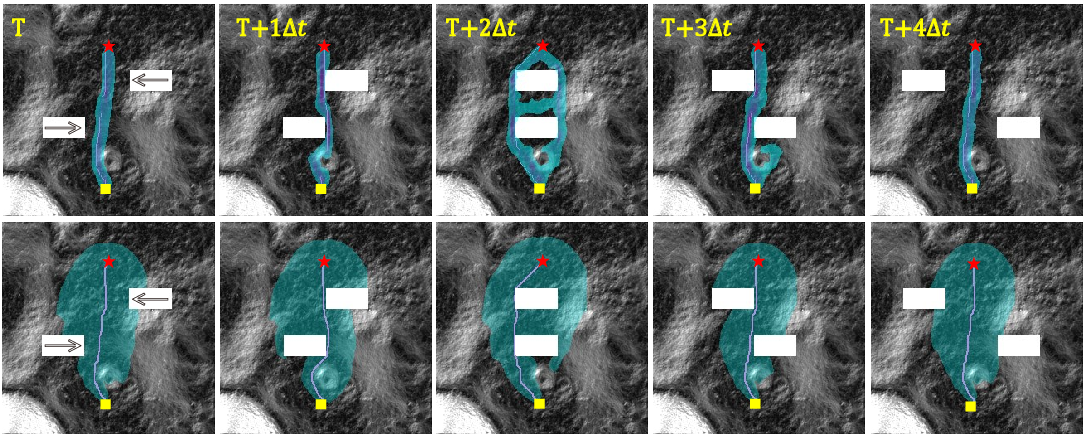}
    \caption{
        The figures illustrate the map changes from time $T$ to $T + 4\Delta t$, with the white rectangular obstacles at the top and bottom representing the added areas with a traversal cost of 1. The upper obstacle moves to the left, and the lower obstacle moves to the right. The first column shows the heuristic region changes (colored in cyan) output by the NNPP model for the five time steps with fixed start and goal points. The second column shows the graphing region (colored in cyan) for path planning using the ${A^\star}$ algorithm under the same conditions.
        }
    \label{fig:re_eps15}}
\end{figure}

\textcolor{revision}
{${D^\star}$ and ${A^\star}$ are considered fundamental algorithms in the field of path planning, with ${D^\star}$ being more commonly employed in dynamic environments. We believe that integrating the heuristic region computed by the NNPP model into either the ${A^\star}$ or ${D^\star}$ algorithm can significantly enhance performance.
The path planning algorithm generates a trajectory based on the current map and start and end points. The agent then navigates along this path until reaching the destination. If the map remains static, replanning is unnecessary. This scenario represents a static environment.
When the map changes during path execution, causing the original path to become suboptimal or obstructed, the algorithm replans the path based on the updated map and the current start and end points.
Incremental search algorithms like ${D^\star}$ efficiently generate new paths by utilizing information from previous iterations, adapting to dynamic environments without replanning from scratch.
}

\begin{figure}[htb]
    \textcolor{revision}{
    \centering
    \includegraphics[width=10 cm]{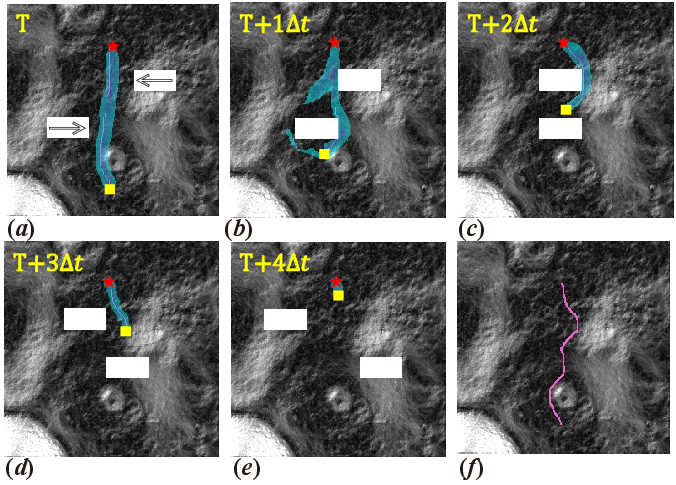}
    \caption{
        (a) to (e) illustrate the map changes at five different time steps. At each time step, the NNPP model is invoked to recompute the heuristic region for the current start and goal positions. (f) depicts the actual path traversed from the initial start to goal, considering the dynamic changes in the environment.
        }
    \label{fig:re_eps16}}
\end{figure}


\textcolor{revision}
{In this paper, we aim to demonstrate that the NNPP model, when combined with the foundation algorithm, can also achieve significant performance improvements for dynamic path planning problems.
}

\begin{table}[t]
    \textcolor{revision}{
    \centering
    \caption{Path planning results for four dynamic time steps. AT represents path planning time, CC represents cumulative traverse cost, and ↑ indicates that the larger the value, the better the performance.}
    \label{tab:tab3}
    \resizebox{\textwidth}{!}{
    \begin{tabular}{>{\hspace{0pt}}m{0.138\linewidth}|>{\centering\hspace{0pt}}m{0.087\linewidth}>{\centering\hspace{0pt}}m{0.108\linewidth}|>{\centering\hspace{0pt}}m{0.087\linewidth}>{\centering\hspace{0pt}}m{0.108\linewidth}|>{\centering\hspace{0pt}}m{0.087\linewidth}>{\centering\hspace{0pt}}m{0.108\linewidth}|>{\centering\hspace{0pt}}m{0.087\linewidth}>{\centering\arraybackslash\hspace{0pt}}m{0.108\linewidth}} 
    \hline\hline
    \multicolumn{1}{>{\Centering\hspace{0pt}}m{0.138\linewidth}|}{\multirow{2}{1\linewidth}{\hspace{0pt}\Centering{}Frames}} & \multicolumn{2}{>{\Centering\hspace{0pt}}m{0.195\linewidth}|}{${A^\star}$}         & \multicolumn{2}{>{\Centering\hspace{0pt}}m{0.195\linewidth}|}{${A^\star NN}$}       & \multicolumn{2}{>{\Centering\hspace{0pt}}m{0.195\linewidth}|}{${D^\star}$}         & \multicolumn{2}{>{\Centering\hspace{0pt}}m{0.195\linewidth}}{${D^\star NN}$}         \\ 
    \cline{2-9}
    \multicolumn{1}{>{\Centering\hspace{0pt}}m{0.138\linewidth}|}{}                                                              & AT↓\textcolor[rgb]{0,0.439,0.753}{} & CC↓\textcolor[rgb]{0,0.439,0.753}{} & AT↓\textcolor[rgb]{0,0.439,0.753}{} & CC↓\textcolor[rgb]{0,0.439,0.753}{} & AT↓\textcolor[rgb]{0,0.439,0.753}{} & CC↓\textcolor[rgb]{0,0.439,0.753}{} & AT↓\textcolor[rgb]{0,0.439,0.753}{} & CC↓\textcolor[rgb]{0,0.439,0.753}{}  \\ 
    \hline
    $T$                                                                                                                            & 1.77                                & 20.61                               & 0.42                                & 20.61                               & 1.73                                & 20.61                               & 0.42                                & 20.61                                \\
    $T+\Delta t$                                                                                                                            & 0.77                                & 15.81                               & 0.25                                & 17.18                               & 0.62                                & 15.81                               & 0.20                                & 16.05                                \\
    $T+2\Delta t$   \textcolor[rgb]{0,0.439,0.753}{}                                                                                         & 0.33                                & 10.62                               & 0.17                                & 11.62                               & 0.24                                & 10.62                               & 0.17                                & 10.62                                \\
    $T+3\Delta t$   \textcolor[rgb]{0,0.439,0.753}{}                                                                                         & 0.17                                & 6.82                                & 0.10                                & 6.82                                & 0.09                                & 6.82                                & 0.11                                & 6.98                                 \\ 
    \hdashline
    all                                                                                                                          & 3.04                                & 53.86                               & 0.94                                & 56.23                               & 2.58                                & 53.86                               & 0.90                                & 54.26                                \\
    \hline\hline
    \end{tabular}}}
\end{table}

\textcolor{revision}
{
To illustrate the effectiveness of our approach, we conducted the following experiment. As shown in Figure \ref{fig:re_eps16}, a path was planned from the yellow rectangular start point to the red star-shaped goal point. Assuming the agent could perfectly follow the planned path, it reached the yellow location in (b) after traversing 45 grids. At this point, the environment map changed, and the previously planned path was blocked by an obstacle. Therefore, the path from the current position to the goal was replanned. After another 45 steps, the environment changed again, and the second planned path was blocked. A third planning was performed, resulting in the path shown in (c). After 45 more steps, the environment changed again. While the previously planned path (the path in (c)) could still lead to the goal without collision, its cumulative traversal cost was not optimal. Therefore, the current environment was re-planned to obtain the optimal path shown in (d). The agent followed this path to reach the goal in (e). Figure \ref{fig:re_eps16} (f) shows the actual path traversed from the initial start point to the goal.
}

\textcolor{revision}
{
    By incorporating the heuristic region of the NNPP model into the ${D^\star}$ algorithm's path search at each iteration, we obtain the ${D^\star}$NN method, which employs the same mechanism as the ${A^\star}$NN method.
}

\begin{table}[tbp]
    \textcolor{revision}{
    \centering
    \caption{Path planning results for five map samples are presented. The '-s' flag indicates a static scene, while the '-d' flag indicates a dynamic scene.}
    \label{tab:tab4}
    \resizebox{\textwidth}{!}{
    \begin{tabular}{>{\centering\hspace{0pt}}m{0.135\linewidth}|>{\centering\hspace{0pt}}m{0.081\linewidth}>{\centering\hspace{0pt}}m{0.121\linewidth}:>{\centering\hspace{0pt}}m{0.067\linewidth}>{\centering\hspace{0pt}}m{0.098\linewidth}>{\centering\hspace{0pt}}m{0.079\linewidth}:>{\centering\hspace{0pt}}m{0.175\linewidth}:>{\centering\hspace{0pt}}m{0.098\linewidth}>{\centering\arraybackslash\hspace{0pt}}m{0.144\linewidth}} 
    \hline\hline
    \multirow{2}{1\linewidth}{\hspace{0pt}\Centering{}scenarios} & \multicolumn{2}{>{\Centering\hspace{0pt}}m{0.202\linewidth}:}{${A^\star}$} & \multicolumn{3}{>{\Centering\hspace{0pt}}m{0.244\linewidth}:}{${A^\star NN}$} & $D^\star$ & \multicolumn{2}{>{\Centering\hspace{0pt}}m{0.242\linewidth}}{${D^\star NN}$}  \\ 
    \cline{2-9}
                                                                     & AT↓  & CC↓                                                                 & AT↓  & CC↓    & MM↑                                                           & AT↓       & AT↓  & CC↓                                                                    \\ 
    \hline
    map1-s                                                           & 2.37 & 120.51                                                              & 0.85 & 122.87 & 0.05                                                          & –         & –    & –                                                                      \\
    map1-d                                                           & 4.13 & 143.87                                                              & 1.46 & 146.75 & 0.05                                                          & 3.08      & 1.13 & 146.57                                                                 \\ 
    \hdashline[1pt/1pt]
    map2-s                                                           & 2.48 & 128.71                                                              & 0.94 & 128.86 & 0.06                                                          & –         & –    & –                                                                      \\
    map2-d                                                           & 3.97 & 161.23                                                              & 1.33 & 164.41 & 0.05                                                          & 2.85      & 1.11 & 164.52                                                                 \\ 
    \hdashline[1pt/1pt]
    map3-s                                                           & 2.31 & 103.15                                                              & 0.67 & 103.42 & 0.09                                                          & –         & –    & –                                                                      \\
    map3-d                                                           & 3.65 & 125.24                                                              & 1.16 & 126.64 & 0.05                                                          & 2.74      & 1.09 & 126.81                                                                 \\ 
    \hdashline[1pt/1pt]
    map4-s                                                           & 2.22 & 92.24                                                               & 0.82 & 93.31  & 0.06                                                          & –         & –    & –                                                                      \\
    map4-d                                                           & 3.72 & 136.82                                                              & 1.24 & 139.55 & 0.07                                                          & 2.89      & 1.12 & 139.55                                                                 \\ 
    \hdashline[1pt/1pt]
    map5-s                                                           & 2.20 & 112.03                                                              & 0.63 & 112.03 & 0.11                                                          & –         & –    & –                                                                      \\
    map5-d                                                           & 3.58 & 134.42                                                              & 1.12 & 136.10 & 0.07                                                          & 2.64      & 0.98 & 135.97                                                                 \\
    \hline\hline
    \end{tabular}}}
\end{table}

\textcolor{revision}
{The experimental data corresponding to the specific timestamps in Figure \ref{fig:re_eps16} are summarized in Table.\ref{tab:tab3} including the time consumption and accumulated cost ($CC$) values for path planning using the ${A^\star}$, ${A^\star NN}$, ${D^\star}$, and ${D^\star NN}$ algorithms. For the initial path planning, ${A^\star}$ and ${D^\star}$ algorithms exhibit similar computational costs due to their underlying principles.
While ${A^\star}$ and ${D^\star}$ guarantee optimal paths, ${D^\star}$ leverages past information to significantly reduce computation time in subsequent planning tasks. This efficiency is further enhanced by ${D^\star NN}$, which achieves a roughly $2.8 \times$speedup compared to ${D^\star}$, as shown in the Table.\ref{tab:tab3}.
}

\textcolor{revision}
{
    At time $T + 3\Delta t$, ${D^\star NN}$ takes 0.11s to compute the path, while ${D^\star}$ algorithm takes only 0.09s. This difference arises from ${D^\star NN}$'s two-step computation process: first, it calls a pre-trained NNPP model to predict the heuristic region, and then it employs a foundation algorithm to plan the optimal path within the reduced area. As illustrated in Figure \ref{fig:re_eps9}, NNPP's inference time averages around 0.06s. Consequently, when the start and end points are close together, NNPP may fall behind foundation algorithms. However, in most task scenarios, NNPP consistently delivers significant performance improvements.
}

\textcolor{revision}
{
    We designed dynamic scenarios similar to Figure \ref{fig:re_eps16} for the five map samples in Figure \ref{fig:re_eps12}. Each map sample was calculated five times and averaged to obtain Table.\ref{tab:tab4}.
    Based on the information presented in Table.\ref{tab:tab4}, we can draw the following four conclusions:
    \begin{itemize}
        \item In static scenarios across five map samples, the ${A^\star NN}$/${A^\star}$ method averaged 0.78s/2.32s, while in dynamic scenarios, the average time for both methods was 1.26s/3.81s. Overall, NNPP achieved a 3$\times$speedup over the ${A^\star}$ algorithm in both static and dynamic scenarios.
        \item In dynamic scenarios, the average time consumption of the ${D^\star}$ algorithm is 2.84 seconds. This demonstrates that the acceleration contribution of NNPP on ${A^\star}$ in dynamic scenarios even exceeds that of the ${D^\star}$ algorithm without NNPP acceleration.
        \item In dynamic scenarios, the average time consumption of the ${D^\star NN}$ method is 1.08s, which shows that NNPP can also bring about a performance improvement of nearly 2.5$\times$speedup for the ${D^\star}$ algorithm.
        \item The NNPP method is expected to result in a 1\%-2\% decrease in path quality.
    \end{itemize}
}

\section{CONCLUSION AND FUTUREWORK}
\textcolor{revision}
{This paper proposes NNPP, a learning-based model for computing heuristic regions, which further accelerates optimal path planning on uneven terrain.}
Drawing inspiration from real-time semantic segmentation models, the study adopts a channel-wise separation of maps, starting points, and destination points specifically for the path planning scenario. Furthermore, the investigation explores the impact of different positional encoding methods on the predictive performance of the model.
The final simulation experiments demonstrate that the proposed method in this paper significantly reduces the computation time required for searching the optimal path in an elevation map compared to the traditional ${A^\star}$ method. Without the need for GPU acceleration, the model achieves an inference speed of 17 FPS, enabling CPU-based real-time path planning in computationally constrained vehicular systems operating in dynamic scenarios. The inference time can be observed in Figure \ref{fig:re_eps9}.

Future work will focus on three aspects. Firstly, efforts will be directed towards enhancing the precision of the model's output in identifying optimal regions. Secondly, the development of model-guided path reconstruction methods will be pursued. As the model output provides a region rather than a specific path, subsequent steps of the algorithm will need to combine traditional methods such as ${A^\star}$ and RRT to reconstruct a feasible path from the starting point to the destination. This leads to the third area of future work, which involves designing an end-to-end network architecture that directly outputs viable paths.

\bibliographystyle{elsarticle-num} 
\bibliography{nnpp_reference_revised}

\clearpage 

\end{document}